\pdfoutput=1

\documentclass[11pt]{article}

\usepackage[final]{acl}
\usepackage{makecell}
\usepackage{times}
\usepackage{latexsym}

\usepackage[T1]{fontenc}

\usepackage[utf8]{inputenc}

\usepackage{microtype}

\usepackage{inconsolata}

\usepackage{graphicx}
\usepackage{listings}

\usepackage{amsmath} 
\usepackage{color}
\usepackage{booktabs}
\usepackage{textcomp}
\usepackage[table]{xcolor}
\usepackage{multirow} 
\newcommand{\start}[1]{\smallskip\vspace{0.5mm}\noindent{{\bf #1}}}

\title{SLM Meets LLM: Balancing Latency, Interpretability and Consistency in Hallucination Detection}

\author{Mengya (Mia) Hu\footnotemark[1],\qquad Rui Xu\footnotemark[1], \qquad Deren Lei\footnotemark[1], \qquad Yaxi Li\footnotemark[1],\qquad Mingyu Wang,\\ \textbf{Emily Ching,\qquad Eslam Kamal,\qquad Alex Deng}\\
Microsoft Responsible AI\\
\{humia, rxu, derenlei, yaxi.li, mwang, yuetc, eskam, alex.deng\}@microsoft.com\\
}

\begin{document}


\newcommand\YAMLcolonstyle{\color{red}\mdseries}
\newcommand\YAMLkeystyle{\color{black}\bfseries}
\newcommand\YAMLvaluestyle{\color{blue}\mdseries}

\makeatletter

\newcommand\language@yaml{yaml}

\expandafter\expandafter\expandafter\lstdefinelanguage
\expandafter{\language@yaml}
{
  basicstyle=\ttfamily,
  backgroundcolor=\color{lightgray},
  frame=single,
  breaklines=true,
  language=,
  keywordstyle=\color{blue},
  commentstyle=\color{green},
  stringstyle=\color{red},
  showstringspaces=false
  basicstyle=\YAMLkeystyle,                                 
  sensitive=false,
  comment=[l]{\#},
  morecomment=[s]{<<}{>>},
  commentstyle=\color{purple}\ttfamily,
  stringstyle=\YAMLvaluestyle\ttfamily,
  moredelim=[l][\color{orange}]{\&},
  moredelim=[l][\color{magenta}]{*},
  morestring=[b]",
  literate =    {---}{{\ProcessThreeDashes}}3
                {\ -\ }{{\mdseries\ -\ }}3,
}

\lst@AddToHook{EveryLine}{\ifx\lst@language\language@yaml\YAMLkeystyle\fi}
\makeatother

\newcommand\ProcessThreeDashes{\llap{\color{cyan}\mdseries-{-}-}}

\maketitle

\renewcommand{\thefootnote}{\fnsymbol{footnote}}
\footnotetext[1]{Equal contributions.}

\footnotetext[2]{This is a preprint of an article that is under review.}

\renewcommand{\thefootnote}{\arabic{footnote}}

\begin{abstract}
Large language models (LLMs) are highly capable but face latency challenges in real-time applications, such as conducting online hallucination detection. To overcome this issue, we propose a novel framework that leverages a small language model  (SLM)  classifier for initial detection, followed by a LLM as constrained reasoner to generate detailed explanations for detected hallucinated content. This study optimizes the real-time interpretable hallucination detection by introducing effective prompting techniques that align LLM-generated explanations with SLM decisions. Empirical experiment results demonstrate its effectiveness, thereby enhancing the overall  user experience.\footnote{\small \url{https://github.com/microsoft/ConstrainedReasoner}}
\end{abstract}

\section{Introduction}
\label{introduction}

\begin{figure*}[ht]
  \centering
  \includegraphics[width=2\columnwidth]{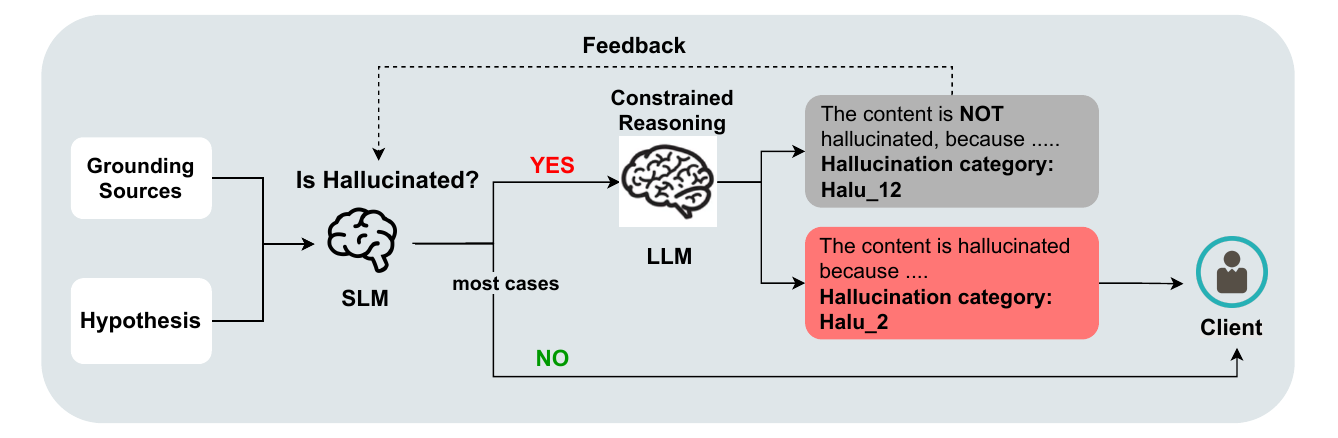}
  \caption{\textbf{Hallucination detection with LLM as constrained reasoner}: Grounding sources and hypothesis pairs are input into a SLM classifier. In most cases, if no hallucination is detected, the no hallucination decision will be returned to the client directly. However, if a hallucination is detected by SLM, an LLM-based constrained reasoner is employed to interpret the SLM's decision. If the reasoner's analysis aligns with the initial hallucination detection, this information, along with the original hypothesis, is relayed to the client.  Otherwise, the potentially problematic hypothesis is filtered out or used as valuable feedback to further refine and improve the upstream SLM.}~\label{fig:hd_flow}
\end{figure*}

Despite Large Language Models (LLMs) having impressive capabilities~\cite{zhou2020detecting, wang2021sketch, wang2020towards, pagnoni2021understanding, dziri2021evaluating}, they are prone to hallucinations—responses that are ungrounded from the source~\cite{rashkin2021increasing, maynezetal2020faithfulness, nanetal2021entity,liu2023lost, Shi2023LargeLM, wei2022chain}-undermining their reliability and making hallucination detection critical~\cite{kaddour2023challenges, pal2023med}.


Conventional hallucination detection methods, such as classification~\cite{kryscinski2019evaluating, zhou2020detecting,zha2023alignscore} or ranking \cite{falke2019ranking} models, have been effective in their domains but often lack interpretability, which is an essential for user trust and mitigation~\cite{rudin2022interpretable}. Given the recent widespread adoption of LLMs, researchers have explored using LLMs for hallucination detection~\cite{lei2023chain, lin2021truthfulqa, min2023factscore, mundler2023self}, utilizing techniques like chain-of-thought reasoning~\cite{ marasovic2021few, kunz2024properties, turpin2024language, shen2023misleading}, or finetuning an autonomous detection agent at Billion-parameter size~\cite{cheng2024small}, or checking consistency of different LLM responses per question~\cite{manakul-etal-2023-selfcheckgpt}. While LLM-based methods provide interpretability, they introduce latency challenges, due to their enormous size and the computational overhead of processing long source texts~\cite{becker2024text, jiang2024longrag}. This creates a major challenge for latency-sensitive real-time applications\footnote{\small \url{https://learn.microsoft.com/en-us/azure/ai-services/content-safety/concepts/groundedness}}. 

We propose a novel workflow to address this challenge by balancing latency and interpretability. Our approach combines a small classification model, which in our case is a small language model (SLM), for initial hallucination detection. A downstream LLM module, termed a "constrained reasoner," then explains the detected hallucinations. This process is illustrated in Figure~\ref{fig:hd_flow}. Considering the relatively infrequent occurrence of hallucinations in practical use~\cite{cao2021hallucinated, wu2023ragtruth, gu2020generating}, the average time cost of using LLMs solely for reasoning on hallucinated texts is manageable. Additionally, this approach leverages the pre-existing reasoning and explanation capabilities of LLMs\cite{ mccoy2023embers}, obviating the need for substantial domain-specific data and significant computational cost on fine-tuning. 

Conventional studies have employed LLMs as end-to-end solutions \cite{sobania2022choose, goyal2022news}. More recently, \citeauthor{shi2024llm} explored the ability of LLMs to explain small classifiers through their latent features, showing promising results on non-reasoning tasks. In this study, we propose a novel framework to effectively apply this approach to hallucination detection.

A potential issue for combining SLM and LLM is the inconsistency between the SLM’s decisions and the LLM’s explanations. Even self-rationalization models, where explanations are generated alongside primary outputs~\cite{wiegreffe2021reframing}, can produce explanations that do not align with the prediction~\cite{ye2022unreliability}. In this study, we focus on addressing such issue in our proposed two-stage hallucination detection framework. Additionally, we analyze LLM reasonings in relation to SLM decisions and ground truth labels, highlighting the potential of LLMs as feedback mechanisms for refining detection processes. 

Our contributions are two-fold: first, we introduce constrained reasoner for hallucination detection that balances latency and interpretability; second, we provide a comprehensive analysis of upstream-downstream consistency, offering practical solutions to enhance the alignment between detection and explanation. We demonstrate its effectiveness on multiple open-source datasets. 

\section{Problem Definition}
We denote the \textit{Grounding Source} as $X$ and the model generated \textit{hypotheses} $Y=(y_1, y_2,...,y_n)$. The generation process can be expressed as a function $\mathcal{F}: X \rightarrow Y$, where $\mathcal{F}$ is the text generation model (e.g., summarization model). $y_i$, where $i \in [1,n]$, is hallucinated if conflicts with or cannot be verified against $X$.

To balance latency and interpretability in hallucination detection, we propose a novel two-stage framework: a SLM for hallucination detection followed by a LLM-based reasoning module, termed "constrained reasoner". The upstream detection can be formulated as: $\mathcal{D}: (X, Y) \rightarrow  J$ where $J =(j_1, j_2,...,j_n)$ represents the binary labels decided by the detector $\mathcal{D}$. The subset of response sentences $Y$ detected as hallucinations by $\mathcal{D}$ is denoted as $H = \{y_k \in Y \mid j_k = \text{hallucination}\} = (h_1, ..., h_m)$, where $m\leq n$. Only detected potential hallucinations $H$ are passed to downstream reasoning module.  The constrained reasoner $\mathcal{R}$ provides explanations for hallucinations flagged by upstream, $\mathcal{R}: (X, H) \rightarrow  E$, where $E = (e_1, ..., e_m)$ contains $m$ explanations, each $e_k,\text{where }  k \in [1,m]$ corresponding to a hallucinated sentence $h_k$ detected by $\mathcal{D}$. $\mathcal{R}$ is called constrained reasoner because it operates under the given constraint that $h_i$ is hallucinated, as determined by $\mathcal{D}$. 
 
However, even in self-rationalization models, reasoning results $E$ may not align with detection results $J$ even they are generated together~\cite{wiegreffe2021reframing,ye2022unreliability}. The inconsistency can be more pronounced in the two-stage frame, where explanations are provided post hoc. We define the real intention in explanation $E$ as $S = (s_1,...,s_m)$. Reasons inconsistent with the upstream decision is thus $\{e_k \in E$, where $s_k = \text{non-hallucination}\}$ (as our framework only passes $\mathcal{R}$ the detected hallucinations to explan due to the latency concern). There are three aspects we want to study regarding the consistency of the constrained reasoner $\mathcal{R}$:

\start{Inconsistency Identification} We design a flagging mechanism to ask LLM-based $\mathcal{R}$ to signal when it judges the hypothesis as non-hallucination and thus unable to provide explanation why the hypothesis is hallucinated. Therefore,  $e_k$ is semi-structured consisting of a free-text reason $t_k$ and a flag $\hat{s}_k$ indicating whether $\mathcal{R}$ thinks the text is hallucination. Formally, $e_k = ( t_k, \hat{s}_k )$. We conduct human evaluation, by asking annotators to careful read $t_k$ and mark $s_k$ whether the reason is explaining the hypothesis is hallucination. Then, we measure effectiveness of the flagging mechanism.

\start{Inconsistency Filtering} The simplest mitigation for inconsistent reasonings is to filter them out. We assess the reduction of inconsistencies after filtering flagged explanations, i.e. ones with $\hat{s}_k = \text{non-hallucination}$. We compare the remaining true inconsistency rates, i.e. the rate of $s_k = \text{non-hallucination}$ as baseline.

\start{Reasoning Feedback} The ground truth label for each $y_i$ is $g_i$, but in practice, $j_i$ may differ from $g_i$ due to SLM imperfections. We explore the potential of $\mathcal{R}$ as a feedback mechanism to improve $\mathcal{D}$. We compare the flagged inconsistencies,  $\hat{s}_k$, against the ground truth $g_k$ to assess $\mathcal{R}$'s performance in identifying non-hallucinations. 

\section{Experiment}
  

Our experiment is designed to study the consistency of reasoning within the proposed hallucination detection framework and effective approaches to filter inconsistencies. Additionally, we explore the potential of LLMs as feedback mechanisms for refining the detection process. We employ GPT4-turbo as $\mathcal{R}$ to elucidate the rationale behind hallucination determinations, using the temperature of 0 and top-p of 0.6. The experiments are conducted across four datasets: \textbf{NHNET} \cite{shen2023misleading}, \textbf{FEVER} \cite{Thorne18Fever}, \textbf{HaluQA} and \textbf{HaluSum} \cite{HaluEval}. We use complete test set of NHNet. Due to the size of rest three datasets and GPT resource limitations, we sample 3000 data per dataset for experimentation. 

To simulate an imperfect SLM classifier, we sample both hallucinated and non-hallucinated responses from the datasets, assuming the upstream label as hallucination. Thus the groundtruth hallucinated text are the simulated true positive cases, and the groundtruth non-hallucinated texts are the the simulated false positive cases. The specific ratio of true and false positives from the SLM is irrelevant to our study, as our focus is on the inconsistencies of the constrained reasoner rather than the performance of the detection algorithm. See appendix \ref{sec:appendix_data} for the distribution of hallucinated and non-hallucinated examples in each dataset. Human annotators assess whether each explanation $e_k$ truly explains why a hypothesis is hallucinated or whether it actually justifies that the text should not be considered a hallucination.

\subsection{Methodology}
\begin{table}
  \centering
\begin{tabular}{|l|c|c|}
    \hline
\makecell{\textbf{Approach}}  & \makecell{\textbf{Fallback } \\ \textbf{when unable} \\ \textbf{to explain}} & \makecell{\textbf{Categorize} \\ \textbf{Hallucinations}} \\
    \hline
   \textit{Vanilla}   & {No}        & {No}       \\
      \hline
    \textit{Fallback}    & {Yes}        & {No}      \\
      \hline
    \makecell{\textit{Categorized}}     & {Yes}         & {Yes}         \\
    \hline
  \end{tabular}
  \caption{Difference between the three main approaches.}
  \label{tab:promptdiff}
\end{table}
The experiment focuses on three primary approaches, with their key distinctions summarized in Table~\ref{tab:promptdiff} (The full prompts are provided in Appendix~\ref{sec:appendix_prompts}).

\start{Vanilla approach} simply instructs $\mathcal{R}$ to explain why the text was detected as hallucination by $\mathcal{D}$. It does not address how to handle inconsistency, i.e. disagreements with the upstream decision. As the reasonings are free-text, there is no straightforward mechanism to identify when inconsistencies arise. If contradictory explanations are generated, they will be presented to the user, which can undermine user trust and experience. It is served as a baseline for Inconsistency Filtering comparison.

\start{Fallback approach} introduces a flagging mechanism whereby $\mathcal{R}$ can respond with "UNKNOWN" to indicate $\hat{s}_k = \text{non-hallucination}$ thus it cannot provide a suitable explanation. This flag helps signal potential inconsistencies, enabling developers to address them effectively.

\start{Categorized approach} refines the flagging mechanism by incorporating more granular hallucination categories. These categories are derived from the analysis of real hallucination data. Among those, a specific category $hallu_{12}$ is used to signal inconsistencies where $\hat{s}_k = \text{non-hallucination}$. By exposing the reasoner to these detailed categories, the goal is to enhance $\mathcal{R}$'s understanding of hallucinations and improve its ability to correctly identify true hallucinations.

\section{Result and Discussion}

\start{Inconsistency Identification} Table~\ref{tab:filterresult} illustrates the performance of identifying real inconsistent reasonings using the designed flags. Both methods demonstrate strong precision. However, the \textit{Fallback} approach exhibits poor recall, i.e. often failing to signal inconsistent reasons with the designed "UNKNOWN" flag. In contrast, \textit{Categorized} approach effectively categorized the majority of inconsistent reasonings under the $hallu_{12}$ flag,  making it easier to filter or mitigate them for downstream usage.

\begin{table}[h]
    \centering
    \scalebox{0.85}{
      \begin{tabular}{l l ccc}
          \toprule
          \textbf{Dataset} & \textbf{Approach} & \textbf{Precision} & \textbf{Recall} & \textbf{F1} \\
          \midrule
          \multirow{2}{*}{FEVER} & \textit{Fallback} & 0.997 & 0.212 & 0.350 \\
                                 & \textit{Categorized} & \textbf{1.000} & \textbf{0.997} & \textbf{0.998} \\
          \midrule
          \multirow{2}{*}{NHNET} & \textit{Fallback} & 0.979 & 0.380 & 0.547 \\
                                 & \textit{Categorized} & \textbf{0.998} & \textbf{0.998} & \textbf{0.998} \\
          \midrule
          \multirow{2}{*}{HaluQA} & \textit{Fallback} & 0.962 & 0.418 & 0.583 \\
                                  & \textit{Categorized} & \textbf{1.000} & \textbf{0.998} & \textbf{0.999} \\
          \midrule
          \multirow{2}{*}{HaluSum} & \textit{Fallback} & \textbf{1.000} & 0.077 & 0.143 \\
                                   & \textit{Categorized} & \textbf{1.000} & \textbf{0.999} & \textbf{0.999} \\
          \bottomrule
    \end{tabular}
    }
    \caption{Inconsistency identification performance based on human evaluations. \textit{Categorized} approach achieves close to perfect performance.}
    \label{tab:filterresult}
  \end{table}

\begin{figure}[t]
\begin{center}
\centerline{\includegraphics[width=\columnwidth]{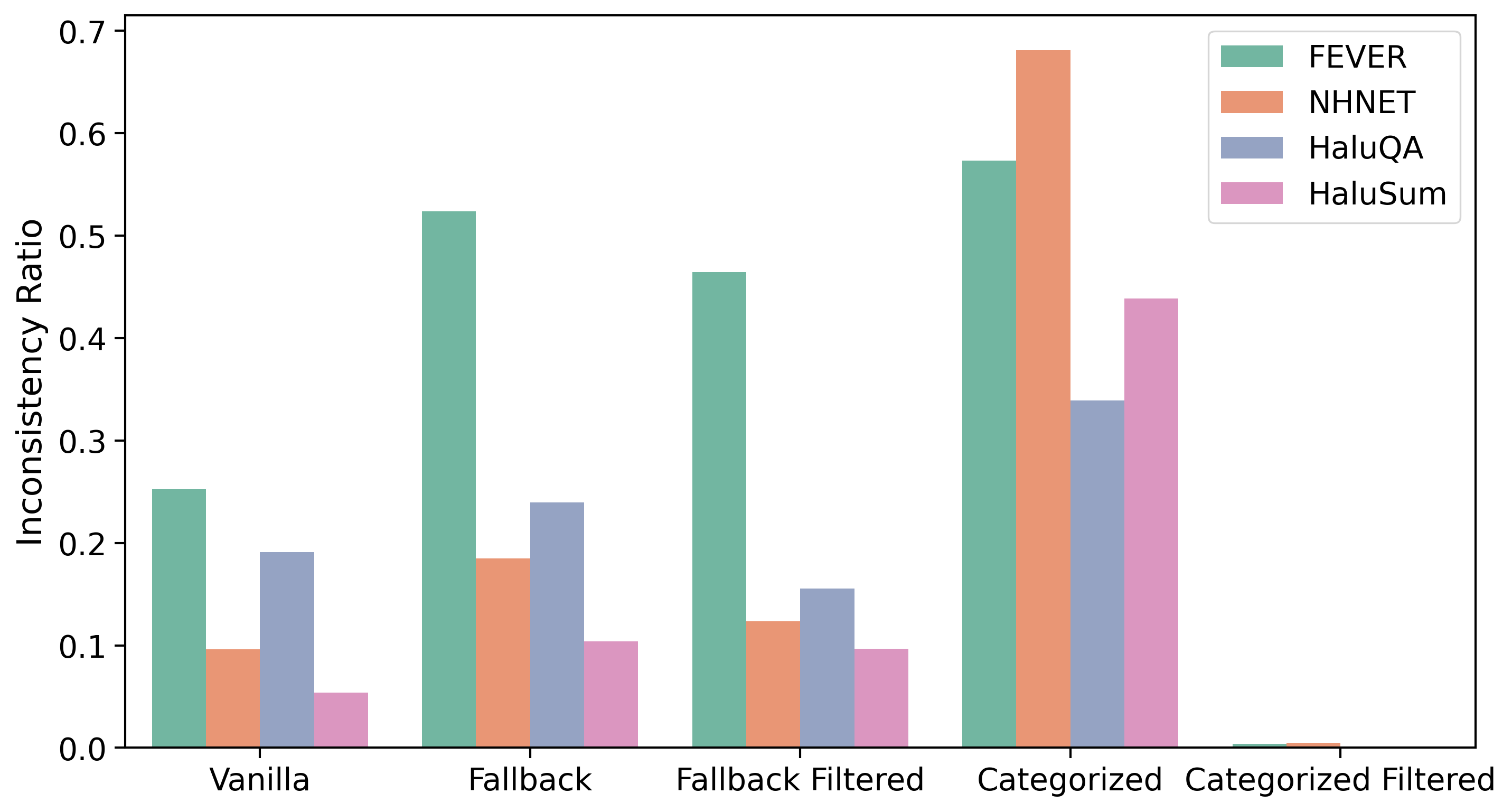}}
\caption{Inconsistency rate comparison: \textit{Categorized} approach consistently outperforms both the \textit{Vanilla} and \textit{Fallback} methods with significant drop in inconsistency after applying filtering.} 
\label{fig:inconc}
\end{center}
\end{figure}

\start{Inconsistency Filtering} Filtering reasonings with the designed flag effectively reduced inconsistencies between the upstream detection and constrained reasoner $\mathcal{R}$, as illustrated in Figure~\ref{fig:inconc}. The \textit{Vanilla} approach, as expected, showed a high inconsistency rate. While the introduction of the "UNKNOWN" category in the \textit{Fallback} approach reduced inconsistencies, its effectiveness was limited by low recall as mentioned above. In contrast, the \textit{Categorized} approach achieved a dramatic reduction across all datasets, with a post-filtering rate as low as $\sim 0.1-1\%$, effectively enhancing the workflow's consistency.


\begin{table}[h]
  \centering
  \scalebox{0.85}{
    \begin{tabular}{l l ccc}
        \toprule
        \textbf{Dataset} & \textbf{Approach} & \textbf{Precision} & \textbf{Recall} & \textbf{F1} \\
        \midrule
        \multirow{2}{*}{FEVER} & \textit{Fallback} & \textbf{1.000} & 0.155 & 0.268 \\
                               & \textit{Categorized} & 0.992 & \textbf{0.778} & \textbf{0.872} \\
        \midrule
        \multirow{2}{*}{NHNET} & \textit{Fallback} & \textbf{0.936} & 0.100 & 0.181 \\
                               & \textit{Categorized} & 0.807 & \textbf{0.820} & \textbf{0.813} \\
        \midrule
        \multirow{2}{*}{HaluQA} & \textit{Fallback} & \textbf{0.968} & 0.201 & 0.333 \\
                                & \textit{Categorized} & 0.901 & \textbf{0.610} & \textbf{0.727} \\
        \midrule
        \multirow{2}{*}{HaluSum} & \textit{Fallback} & \textbf{0.792} & 0.013 & 0.026 \\
                                 & \textit{Categorized} & 0.763 & \textbf{0.669} & \textbf{0.713} \\
        \bottomrule
  \end{tabular}
  }
  \caption{Feedback results based on LLM constrained reasoning and ground truth labels. \textit{Categorized} approach consistently achieves higher recall and F1.}
  \label{tab:feedbackresult}
\end{table}

\start{Reasoning Feedback} As results shown in Table~\ref{tab:feedbackresult}, the \textit{Categorized} approach demonstrated strong potential as feedback mechanism, outperforming the \textit{Fallback} method with high recall. It achieves a macro-average F1 score of 0.781. This indicates its capability to accurately identify false positives from the SLM, making it a promising feedback mechanism for improving the upstream model—an area worth further exploration. The high inconsistency rate observed in the \textit{Categorized} approach before filtering, as shown in Figure~\ref{fig:inconc}, highlights the ability of LLMs like GPT to accurately identify true hallucinations when refined hallucination categories are provided, as indicated by the high F1 in Table~\ref{tab:feedbackresult}. This suggests that LLM can maintain correct judgments without being easily influenced or swayed by specific instructions.

\section{Conclusion}
In this study, we introduce a practical framework for efficient and interpretable hallucination detection by combining SLM for detection with LLM for constrained reasoning. Our \textit{Categorized} prompting strategy with filtering effectively aligns LLM explanations with SLM decisions, empirically proven effective on 4 hallucination and factual consistency datasets. Furthermore, this strategy shows promise as a feedback mechanism for refining SLMs, offering a path toward more robust and adaptive systems. While our experiments focus on real-time interpretable hallucination detection, the insights gained are broadly applicable, shades lights in improving classification decision systems and enhancing SLM capabilities through LLM-based constrained interpretation. 
\bibliography{custom}

\appendix

\section{Appendix}
\label{sec:appendix}
\subsection{Data distribution}
\label{sec:appendix_data}
The distribution of hallucinated and non-hallucinated examples in each dataset is shown in Table~\ref{tab:data}.
\begin{table}[h]
  \centering
  
  \scalebox{0.8}{
\begin{tabular}{|l|c|c|}
    \hline
\makecell{\textbf{Dataset}}  & \makecell{\textbf{Hallucination}} & \makecell{\textbf{Non-Hallucination}} \\
    \hline
   NHNET  & {216}        & {439}       \\
      \hline
    FEVER     & {813}         & {2187}      \\
    \hline
    HaluQA     & {1500}         & {1500}      \\
    \hline
    HaluSum     & {1500}         & {1500}      \\
    \hline
  \end{tabular}
  }
  \caption{\textbf{Dataset statistics}. NHNET we use the complete set. Fever, HaluQA and HaluSum we random sample 3000 data due to their large size.}
  \label{tab:data}
\end{table} 

\subsection{Constrained Reasoner Approaches}
\label{sec:appendix_prompts}
\subsubsection{Vanilla prompt}
Vanilla prompt shown in \ref{fig:basic_prompt} only gives the instruction and few-shot examples to do the downstream reasoning task. However, it does not specify how LLM should deal with the situation where LLM does not follow the upstream decision.
\begin{figure*}[t]
\begin{lstlisting}[language=yaml, basicstyle=\tiny]
 
- role: system
  content: |
        You are a careful proof-reading assistant with great logic thinking and solid english skills for a documentation scribe. Your important task is to provide hallucination reasons: given the <<Source Document>> and some <<Sentences>> that is not supported by the <<Source Document>>, you are expected to give the <<Reason>> why the sentence are not supported.
        If the <<Sentences>> contradict the <<Source Document>>, you should cite the evidence in the <<Source Document>> and specify where the contradiction is.
        If the hallucination is because a small part of the <<Sentences>> is made up/ no information in the <<Source Document>> supports/contradicts the small part of the sentence, please "PARTIAL NEUTRAL" and specify which part is not supported.
        If the hallucination is because the whole <<Sentence>> is made up/ no information in the <<Source Document>> supports/contradicts the sentence, please mark "NEUTRAL" to mark this situation. Please try your best to find the detailed reasons and only use NEUTRAL as your last resort.
        If there are part of the <<Sentence>> contradicts and part of the <<Sentence>> "NEUTRAL", please specify all the reasons. 
        The <<Sentences>> are numbered. You should provide the <<Reason>> in the same order as the original <<Sentences>>.

- role: system
  name: example_user
  content: |
        Let's try it.  
        <<Source Document>>:
        The Academy Awards, also known as the Oscars are awards for artistic and technical merit for the film industry. They are presented annually by the Academy of Motion Picture Arts and Sciences, in recognition of excellence in cinematic achievements as assessed by the Academy's voting membership. The Academy Awards are regarded by many as the most prestigious, significant awards in the entertainment industry in the United States and worldwide. The awards ceremony is always hosted in the US.
        <<End Source Document>>

        <<Sentences need to provide hallucination reasons>>:
        (0). <<Sentence>>: Oscar is presented every other two years.
        (1). <<Sentence>>: Will Smith won the 2022 Oscar.
        (2). <<Sentence>>: The awards ceremony is always hosted in the US in summer.
        <<End Sentences need to provide hallucination reasons>>

- role: system
  name: example_assistant
  content: |
        These are hallucinations because:
        (0). the source reference: "They are presented annually by the Academy of Motion Picture Arts and Sciences", thus it is not presented every other two year. It's contradiction.
        (1). NEUTRAL
        (2). PARTIAL NEUTRAL. The main part of the sentence is correct, but the grounding source did not mention "summer". 

- role: system
  name: example_user
  content: |
        Let's try it again.  
        <<Source Document>>:
        Prompts are how you ask Copilot to do something for you - like creating, summarizing, editing, or transforming. Think about prompting like having a conversation, using plain but clear language and providing context like you would with an assistant.
        Also called prompt engineering, prompting is both an art and a science. To get the best results, you need to structure your prompt in a way that the large language model (LLM) can understand.
        Like any other skill, prompting takes practice to perfect. You won't get there overnight.  
        How to write a good prompt? 
        <<End Source Document>>

        <<Sentences need to provide hallucination reasons>>:
        (0). <<Sentence>>: Give clarity and Context and you will do a good job immediately.
        <<End Sentences need to provide hallucination reasons>>

- role: system
  name: example_assistant
  content: |
        These are hallucinations because:
        (0). The grounding source is a reference and a user question. The "clarity and Context" in the answer sentence is correct, but the "you will do a good job immediately" contradicts the source: "prompting takes practice to perfect. You won't get there overnight."

- role: user
  content: | 

        <<Source Document>>:
        {{transcript}}
        <<End Source Document>>

        <<Sentences need to provide hallucination reasons>>:
        {{sentences}}
        <<End Sentences need to provide hallucination reasons>>

        Give your reason and begin your answer with "These are hallucinations because:\n"
\end{lstlisting}
\caption{\textbf{Vanilla prompt}. }
  \label{fig:basic_prompt}
\end{figure*}

\subsubsection{Fallback Prompt}
Fallback prompt shown in \ref{fig:fallback_prompt} gives LLM an alternative route when it does not agree with the upstream decision and will give inconsistent downstream explanations. 
\begin{figure*}[t]
\begin{lstlisting}[language=yaml, basicstyle=\tiny]
- role: system
  content: |
        You are a careful proof-reading assistant with great logic thinking and solid english skills for a documentation scribe. Your important task is to provide hallucination reasons: given the <<Source Document>> and some <<Sentences>> that is not supported by the <<Source Document>>, you are expected to give the <<Reason>> why the sentence are not supported.
        If the <<Sentences>> contradict the <<Source Document>>, you should cite the evidence in the <<Source Document>> and specify where the contradiction is.
        If the hallucination is because a small part of the <<Sentences>> is made up/ no information in the <<Source Document>> supports/contradicts the small part of the sentence, please "PARTIAL NEUTRAL" and specify which part is not supported.
        If the hallucination is because the whole <<Sentence>> is made up/ no information in the <<Source Document>> supports/contradicts the sentence, please mark "NEUTRAL" to mark this situation. Please try your best to find the detailed reasons and only use NEUTRAL as your last resort.
        If there are part of the <<Sentence>> contradicts and part of the <<Sentence>> "NEUTRAL", please specify all the reasons. 
        The <<Sentences>> are numbered. You should provide the <<Reason>> in the same order as the original <<Sentences>>.
        In very rare case, if you can not find the reason for the hallucination or you think the <<Sentences>> is supported by the <<Source Document>>, please mark 'UNKNOWN'.

- role: system
  name: example_user
  content: |
        Let's try it.  
        <<Source Document>>:
        The Academy Awards, also known as the Oscars are awards for artistic and technical merit for the film industry. They are presented annually by the Academy of Motion Picture Arts and Sciences, in recognition of excellence in cinematic achievements as assessed by the Academy's voting membership. The Academy Awards are regarded by many as the most prestigious, significant awards in the entertainment industry in the United States and worldwide. The awards ceremony is always hosted in the US.
        <<End Source Document>>

        <<Sentences need to provide hallucination reasons>>:
        (0). <<Sentence>>: Oscar is presented every other two years.
        (1). <<Sentence>>: Will Smith won the 2022 Oscar.
        (2). <<Sentence>>: The awards ceremony is always hosted in the US in summer.
        <<End Sentences need to provide hallucination reasons>>

- role: system
  name: example_assistant
  content: |
        These are hallucinations because:
        (0). the source reference: "They are presented annually by the Academy of Motion Picture Arts and Sciences", thus it is not presented every other two year. It's contradiction.
        (1). NEUTRAL
        (2). PARTIAL NEUTRAL. The main part of the sentence is correct, but the grounding source did not mention "summer". 

- role: system
  name: example_user
  content: |
        Let's try it again.  
        <<Source Document>>:
        I have bluntly communicated to him several times that I would like to see him more than once per week for two hours. He said he would try to make more time for me but it has not happened yet (...) I don't smoke weed anymore, and I don't care if he does but I'm hurt that he barely finds time to spend with me. I don't even text or call to ask him to hang out because he'll usually say he's busy working. I know he's busy and throughout the week I'm busy with work too but we rarely talk. He'll probably send a text every so often, but we usually go 4-5 days without talking. (...) Sometimes the one day I do get to hang out with him, he'll cancel if someone calls and I'll be left at dinner/movie theater/wherever alone. I just feel like I'm not wanted.
        <<End Source Document>>

        <<Sentences need to provide hallucination reasons>>:
        (0). <<Sentence>>: I feel like my boyfriend is neglecting me and I'm not sure what to do about it.
        <<End Sentences need to provide hallucination reasons>>

- role: system
  name: example_assistant
  content: |
        These are hallucinations because:
        (0). UNKNOWN. Although original article does not use the word 'neglecting', it a lot of incidence that we may inferred 'neglecting'. Therefore, I can not categorize the reason and need to use UNKNOWN.

- role: system
  name: example_user
  content: |
        Let's try it again.  
        <<Source Document>>:
        Prompts are how you ask Copilot to do something for you - like creating, summarizing, editing, or transforming. Think about prompting like having a conversation, using plain but clear language and providing context like you would with an assistant.
        Also called prompt engineering, prompting is both an art and a science. To get the best results, you need to structure your prompt in a way that the large language model (LLM) can understand.
        Like any other skill, prompting takes practice to perfect. You won't get there overnight.  
        How to write a good prompt? 
        <<End Source Document>>

        <<Sentences need to provide hallucination reasons>>:
        (0). <<Sentence>>: Give clarity and Context and you will do a good job immediately.
        <<End Sentences need to provide hallucination reasons>>
\end{lstlisting}   
\end{figure*}

\begin{figure*}[t]
\begin{lstlisting}[language=yaml, basicstyle=\tiny]
- role: system
  name: example_assistant
  content: |
        These are hallucinations because:
        (0). The grounding source is a reference and a user question. The "clarity and Context" in the answer sentence is correct, but the "you will do a good job immediately" contradicts the source: "prompting takes practice to perfect. You won't get there overnight."

- role: user
  content: | 

        <<Source Document>>:
        {{transcript}}
        <<End Source Document>>

        <<Sentences need to provide hallucination reasons>>:
        {{sentences}}
        <<End Sentences need to provide hallucination reasons>>

        Give your reason and begin your answer with "These are hallucinations because:\n"
\end{lstlisting}   
\caption{\textbf{Fallback Prompt}. }
  \label{fig:fallback_prompt}
\end{figure*}

\subsubsection{Categorized Prompt}
Categorized prompt shown in \ref{fig:categorized_prompt} gives LLM an alternative route when it does not agree with the upstream decision and will give inconsistent downstream explanations. Moreover, this prompt asks the LLM to categorize the reasons when LLM agrees with the upstream decision as an extra confirmation.
\begin{figure*}[t]
\begin{lstlisting}[language=yaml, basicstyle=\tiny]
- role: system
  content: |
        You are a careful proof-reading assistant with great logic thinking and solid english skills for a documentation scribe. Your important task is to provide hallucination reason categories: given the <<Source Document>> and some <<Sentences>> that is not supported by the <<Source Document>>, i.e. the <<Sentences>> are hallucinated, you are expected to give the <<Hallucination Categories>> why the sentence are not supported.
        <<Hallucination Categories>> and their definitions are:
        Hallu_1. Missing from grounding sources: At least one factual claim in response sentence is not present in grounding sources and cannot be inferred using basic domain/common knowledge. 
        Hallu_2. Numeric value contradicts with ground source: Numeric value with similar context occurs in the grounding source but the value in response sentence contradicts with the value that occurs in the grounding source.
        Hallu_3. Negative to positive flip: At least factual claim also occurs in the grounding sources but appears as negative in the grounding source and change to positive  in the response sentence  
        Hallu_4. Positive to negative flip: At least factual claim also occurs in the grounding sources but appears as positive in the grounding source and change to negative  in the response sentence 
        Hallu_5. Entity grouped wrong: Response sentence categorized an entity which occurs in the grounding source, incorrectly 
        Hallu_6. Url contradicts with ground source: Url with similar context occurs in the grounding source but the url in response sentence contradicts with the url in the grounding source  
        Hallu_7. Missing information changes meaning: A part of the information in the grounding source is missing from the response. This changes the meaning of the fact or entity stated. 
        Hallu_8. Claim contradicts with grounding source: Factual claim contradicts with the information in the ground source (but the claim is not a numeric value or url) 
        Hallu_9. Pronoun contradicts with grounding source: Incorrect pronouns cause overwise correct sentence to become incorrect. (EG OfficeSum I: 22, S: 3: "He believes the high number of twins could be genetic." The grounding sources state that some experts believe it could be genetic. The sentence is correct except for the pronoun he.) 
        Hallu_10. Template sentence fabricated hallucination: Some summaries use templates. These templates elicit answers even when the answer is not mentioned in the grounding source. For example: "patient is accompanied by adult female" is a common template hallucination. 
        Hallu_11. Source misspelt/unintelligible.
        Hallu_12. All others. 
        Try your best to classify the reasons in to the first 11 categories. In very rare case, if you can not find the reason for the hallucination or you think the <<Sentences>> is supported by the <<Source Document>>, please use 'Hallu_12'. That means you can only provide the above 12 categories! 
        The <<Sentences>> are numbered. You should provide the <<Hallucination Categories>> in the same order as the original <<Sentences>>.
\end{lstlisting}   
\end{figure*}

\begin{figure*}[ht]
\begin{lstlisting}[language=yaml, basicstyle=\tiny]        
- role: system
  name: example_user
  content: |
        Let's try it.  
        <<Source Document>>:
        The Academy Awards, also known as the Oscars are awards for artistic and technical merit for the film industry. They are presented annually by the Academy of Motion Picture Arts and Sciences, in recognition of excellence in cinematic achievements as assessed by the Academy's voting membership. The Academy Awards are regarded by many as the most prestigious, significant awards in the entertainment industry in the United States and worldwide. The awards ceremony is always hosted in the US. A promotion about Oscar begins at 3:300PM June 29, 2012.
        <<End Source Document>>

        <<Sentences need to provide hallucination reasons>>:
        (0). <<Sentence>>: Oscar is presented every other two years.
        (1). <<Sentence>>: Will Smith won the 2022 Oscar.
        (2). <<Sentence>>: The awards ceremony is always hosted in the US in summer.
        (3). <<Sentence>>: An event begins at 3:00 PM and ends at 3:00 pm EST on July 5th, 2012. 
        <<End Sentences need to provide hallucination reasons>>

- role: system
  name: example_assistant
  content: |
        These are hallucinations because:
        (0). Hallu_2. The source reference: "They are presented annually by the Academy of Motion Picture Arts and Sciences", thus it is not presented every other two year, i.e. contradiction in numeric value.
        (1). Hallu_1. The source reference did not mention Will Smith won the 2022 Oscar.
        (2). Hallu_1. The main part of the sentence is correct, but the grounding source did not mention "summer". 
        (3). Hallu_7. The original article mentioned "begins at 3:300PM June 29, 2012". Date is missed in sentence, as the result the meaning becomes begin time is 3:00PM July 5th, 2012 which is wrong. 

- role: system
  name: example_user
  content: |
        Let's try it again.  
        <<Source Document>>:
        Prompts are how you ask Copilot to do something for you - like creating, summarizing, editing, or transforming. Think about prompting like having a conversation, using plain but clear language and providing context like you would with an assistant.
        Also called prompt engineering, prompting is both an art and a science. To get the best results, you need to structure your prompt in a way that the large language model (LLM) can understand.
        Like any other skill, prompting takes practice to perfect. You won't get there overnight.  
        How to write a good prompt? 
        <<End Source Document>>

        <<Sentences need to provide hallucination reasons>>:
        (0). <<Sentence>>: Give clarity and Context and you will do a good job immediately.
        <<End Sentences need to provide hallucination reasons>>

- role: system
  name: example_assistant
  content: |
        These are hallucinations because:
        (0). Hallu_8. The grounding source is a reference and a user question. The "clarity and Context" in the answer sentence is correct, but the "you will do a good job immediately" contradicts the source: "prompting takes practice to perfect. You won't get there overnight."

- role: system
  name: example_user
  content: |
        Let's try it again.  
        <<Source Document>>:
        I have bluntly communicated to him several times that I would like to see him more than once per week for two hours. He said he would try to make more time for me but it has not happened yet (...) I don't smoke weed anymore, and I don't care if he does but I'm hurt that he barely finds time to spend with me. I don't even text or call to ask him to hang out because he'll usually say he's busy working. I know he's busy and throughout the week I'm busy with work too but we rarely talk. He'll probably send a text every so often, but we usually go 4-5 days without talking. (...) Sometimes the one day I do get to hang out with him, he'll cancel if someone calls and I'll be left at dinner/movie theater/wherever alone. I just feel like I'm not wanted.
        <<End Source Document>>

        <<Sentences need to provide hallucination reasons>>:
        (0). <<Sentence>>: I feel like my boyfriend is neglecting me and I'm not sure what to do about it.
        <<End Sentences need to provide hallucination reasons>>

- role: system
  name: example_assistant
  content: |
        These are hallucinations because:
        (0). Hallu_12. Although original article does not use the word 'neglecting', it a lot of incidence that we may inferred 'neglecting'. Therefore, I can not categorize the reason into the first 11, and need to use H12.

- role: user
  content: | 

        <<Source Document>>:
        {{transcript}}
        <<End Source Document>>

        <<Sentences need to provide hallucination reasons>>:
        {{sentences}}
        <<End Sentences need to provide hallucination reasons>>

        Give your reason and begin your answer with "These are hallucinations because:\n"
\end{lstlisting}
\caption{\textbf{Categorized Prompt}. }
  \label{fig:categorized_prompt}
\end{figure*}

\end{document}